# A Data Driven Method for Multi-step Prediction of Ship Roll Motion in High Sea States

Dan Zhang[1,2#], Xi Zhou[1,2#], Zi-Hao Wang[2,3*], Yan Peng[1,2,3] and Shao-Rong Xie[4]

[1]School of Mechatronic Engineering and Automation, Shanghai University, Shanghai, 200444, China

[2]Engineering Research Center of Unmanned Intelligent Marine Equipment, Ministry of Education, Shanghai, 200444, China.

[3]School of Artificial Intelligence, Shanghai University, Shanghai, 200444, China

[4]School of Computer Engineering and Science, Shanghai University, Shanghai, 200444, China

Ship's roll motion in high sea state has large amplitude and nonlinear dynamics, and its prediction is significant for the operability, safety and survivability. This paper presents a novel data-driven methodology to provide multi-step prediction of the ship's roll motion in high sea states. A hybrid neural network is proposed that combines long short-term memory (LSTM) and convolutional neural network (CNN) in parallel. The motivation is to extract the nonlinear dynamics characteristics and the hydrodynamic memory information through the advantage of CNN and LSTM, respectively. For the feature selection, the time histories of motion states and wave heights are selected to involve sufficient information. Taken a scaled KCS as the study object, the ship motions in sea state 7 irregular long crested waves are simulated and used for the validation. The results show that at least one period of roll motion can be accurately predicted by using the proposed method. Compared with the single LSTM and CNN method, the proposed method has better performance in the prediction of the amplitude of roll angles. Besides, the comparison results also demonstrate that selecting motion states and wave heights as feature space improves the prediction accuracy, verifying the effectiveness of the proposed method.

[#]These authors contributed equally to this work and should be considered co-first authors
[*]Corresponding author, e-mail address: zihaowang@shu.edu.cn

# 1 Introduction

When a ship encounters high sea states, severe motion can be induced by the wave-dominated environmental disturbances, significantly affecting operability, safety and survivability. The prediction of ship roll motion is critical to motion compensation, which may prevent cargo crash during cargo transfer and improve the aircrafts landing safety on carrier and firing accuracy of shipboard weapons systems (Huang et al. 2018). This prediction information is also helpful to ship captain or autopilot to make proper decisions. However, due to the highly complex interaction of the ship hull with the incoming waves as well as the stochastic character of ocean waves, the accurate prediction of the roll motion in high sea state is still a challenge.

Prediction approaches for ship roll motion can be broadly classified into mechanism method and data-driven method. Mechanism modeling requires an in-depth analysis of the physical mechanism and generally establishes a mathematical model consisting of the damping moment, the restoring moment, the added moment of inertia and the wave disturbance moment. To obtain the model parameters, several methods have been investigated, including model test methods (Diez et al., 2018), empirical methods (Inoue et al., 1981; Newman, 2018; Yasukawa and Yoshimura, 2015), numerical methods and system identification methods (Hou and Zou, 2015; Jiang et al., 2021). Nevertheless, the ship roll motion in high sea states has large amplitude and belongs to typical nonlinear motion. In this case, wave disturbance moments are highly coupled with other moments caused by the ship motion. As a result, the complex expression of hydrodynamic coefficient in the mechanism model requires huge amount of calculation, and fixed hydrodynamic coefficient of the ship under high sea state is difficult to adapt to the complex and changeable ocean environment. Therefore, the conventional hydrodynamic component model is unsuitable for accurate prediction of roll motions in high sea state (Bassler, 2013).

The data-driven modeling method extracts dynamic characteristics from the historical data of states. The classic time series modeling methods optimize the model structure from the priori model set and then identify weight parameters, such as auto regressive (AR) model and auto regressive moving average (ARMA) model (Jiang et al., 2020; Yumori, 1981). Despite their high efficiency, their forecasting of ship roll motion is not effective in rough sea states since the presence of nonlinearities. To express the nonlinear features of ship roll motions, machine learning methods have been investigated, which can map the input features to a high-dimensional space. Yin et al. (2013) combined radial basis function with sliding windows for ship roll motion predictions under moderate sea states. Huang et al. (2018) proposed a structural wavelet neural network for high accurate prediction of roll motion under regular waves. These studies mainly considered the current states as the feature variables.

It is worth noting that the ship roll motion in waves has a significant memory effect (Chung and Bernitsas, 1997; Liu et al., 2020), that is, the hydrodynamic forces are not only related to the current states, but also to the time history of these states. For this issue, recurrent-type neural networks are suitable to be applied, which have shown good performance in extracting the time correlation in time series (Sutskever et al., 2014). Among them, the long short-term memory (LSTM) neural networks overcome the shortcomings of recurrent neural networks (RNN) that are prone to gradient disappearance, and improve the learning ability of long sequences by adding gating units (Hochreiter and Schmidhuber, 1997). Liu et al. (2020) determines the optimal vector space of LSTM by Impulse Response Function and Auto-correlation Function, predicting the roll motion in one step. Tang et al. (2021) adopted a LSTM to predict the time series of ship motion, where the dataset under moderate sea states was simulated by wave energy spectrum and strip theory. Wei et al. (2021) decomposed the sequence of roll motion in moderate sea state with the adaptive empirical wavelet transform method for LSTM model, achieving accurate multi-step prediction of roll motions. In general, LSTM shows good performance in predicting ship rolling motion. The hybrid model is also employed to get more accurate future roll motion. The coupled CNN and LSTM (Zhang et al., 2019) was adopted to tackle the multidimension data from USV sensor so as to predict the one step of roll motion of USV under low sea state. However, the above studies mainly focus on low and moderate sea states, and further research is still needed for the prediction of roll motion in high sea states with large roll angles.

For data-driven approaches, datasets play a key role in model training. But available datasets under high sea state are scarce due to the high cost and navigational safety risks. Numerical techniques based on computational fluid dynamics (CFD) provide an effective method to get relatively realistic simulation of ship motion in high sea state (Diez et al., 2018; Jiao et al., 2019; Serani et al., 2021; Wang et al., 2017). Based on the CFD simulation data, several studies investigated the modeling of roll motion under high sea states.

Besides, with the development of real-time wave measurement techniques based on visual techniques (Bergamasco et al., 2021; Cang et al., 2019) and wave radar techniques (Lyzenga et al., 2015; Wang et al., 2007), the acquisition of wave height is becoming possible and drawn the modelers' attention. Based on the CFD simulation data, several studies investigated the data-driven modeling of roll motion with wave as input in high sea states. Del Águila Ferrandis et al. (2021) performed one-step prediction of DTMB class ship motion under 7th sea state using RNN, Gate Recurrent Unit (GRU) and LSTM based on the CFD simulation dataset, with wave height as the input feature. Xu et al. (2021) predicted the ship's roll and heave motions under simulated irregular beam waves of sea state 7 based on LSTM method, which also utilized only the wave height information at a position in front of the ship. As can be seen, wave height was chosen as the input feature in both studies, which is expected to provide direct information about wave impacts. However, only one step of prediction was tested in these studies, and further research on multi-step prediction is still required.

Until recently, few studies have focused on multi-step prediction of ship roll motion in high sea states. D'Agostino et al. (2021) investigated short-term prediction of ship multi-degree-of-freedom motions in high sea states based on CFD simulations, focusing on comparing the capabilities of RNN, GRU and LSTM. However, the impact of the feature variables on multi-step prediction still needs to be explored. Inspired by the latest studies, this study focuses on the multi-step prediction of ship roll motion in high sea states, and comprehensively investigates the effects of the feature variables.

This paper presents a novel data-driven modeling method for multi-step predicting the ship roll motion in high sea states. A neural network framework named ConvLSTMPNet is proposed to extract the nonlinear dynamics characteristics and the hydrodynamic memory information from wave heights and roll motion state, in which LSTM and CNN are called in parallel as feature extraction modules. Taking the KCS ship model as the research object, the CFD method is utilized to generate the motion data in sea state 7 irregular long-peaked waves with different wave directions. A comparative study on the selection of feature space is conducted, considering the effects of time history of motion states and wave height. In addition, in order to verify the performance of ConvLSTMPNet on multi-steps roll prediction, single LSTM and single CNN model the is adopted as baseline models.

## 2 Problem description

The purpose of this study is to accurately predict the ship roll motion in high sea states over a period of time, which helps captain or autopilot obtain trends over a future horizon and make the proper decisions to enhance operability and safety. In general, the ship roll motion in high sea states has large amplitudes and nonlinear dynamics. In this condition, the damping moment, restoring moment, added moment of inertia and wave disturbance moment have interaction effects, making the assumption of the conventional mechanism models invalid and thus unable to provide accurate predictions. Meanwhile, the interactions among the above-mentioned moments induce stronger nonlinearities. To construct hydrodynamic memory effects of ship roll motion in high sea state, a time series problem is considered and a multi-step prediction model will be developed by a data-driven approach.

Multi-step forecasting models predict a horizon of future values by all available inputs through a sequence-to-sequence architecture. It can be viewed as a modification of the one-step-head forecasting problem. In this study, the multi-step forecasting model of ship roll motion takes the form:

$$(y_{t+1}, y_{t+2}, \ldots, y_{t+p}) = f(y_{t-d+1}, y_{t-d+2}, \ldots, y_t, X_t) \tag{1}$$

where $(y_{t+1}, y_{t+2}, \ldots, y_{t+p})$ is a discrete forecast horizon of roll angle, $(y_{t-d+1}, y_{t-d+2}, \ldots, y_t)$ are the observations of the roll angle over a look-back window $d$ and $X_t = (x_{t-d+1}, x_{t-d+2}, \ldots, x_t)$ are the exogenous inputs over a look-back window *d*. The $f(\cdot)$ is the nonlinear function to be learned. Thus, the forecasting modeling can be described as a supervised learning problem. The key issues of the modeling method are how to determine the feature variables and how to design the modeling architecture to extract the information of nonlinear dynamics and hydrodynamic memory effects, thereby achieving accurate multi-step predictions.

## 3 Methodology

### 3.1 Overview of the methodology

The data-driven methodology mainly includes the design of the learning algorithm and the selection of the feature variables. In addition, due to the specificity of the high sea state conditions, the data scarcity of the training set also needs to be considered.

In terms of learning algorithms, to better extract the nonlinear dynamics and hydrodynamic memory effects of roll motions, a hybrid neural network is proposed that combines long short-term memory (LSTM) and convolutional neural network (CNN) in parallel, exploiting the CNN's ability to extract the multi-variable coupling relationships and the LSTM's ability to extract sequential correlation. As regards the feature variables, both the roll motion states and the wave heights are investigated. To address the data scarcity of the training set, the numerical computation through CFD method is used to provide high-fidelity dataset.

The overall procedure of the methodology is displayed in Fig. 1. Firstly, to address the data scarcity, the CFD method generates the ship motion data in high sea state with different wave directions. Secondly, the sliding window method is adopted to obtain the dataset suitable for the supervised learning. Then, the proposed ConvLSTMPNet combined by CNN and LSTM network in parallel is used to extract nonlinear dynamics and hydrodynamic memory effect information, and the multi-steps roll prediction is interpreted by the fully connected layers.

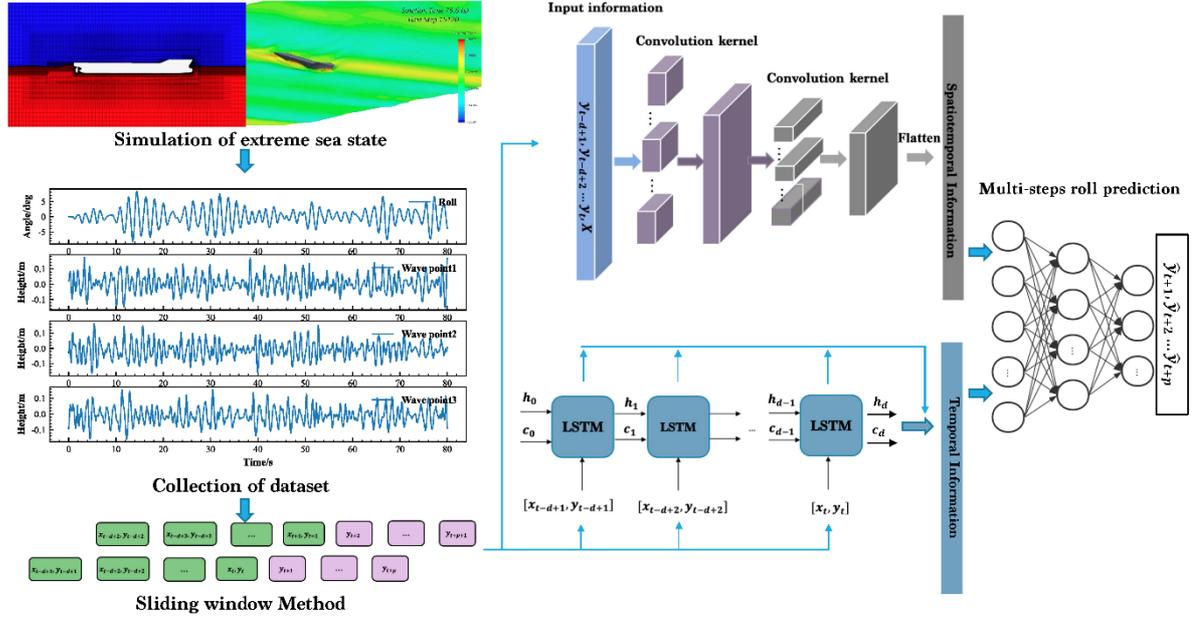

Fig. 1 Overview of the forecasting modeling procedure

3.2 Multi-step forecasting model: ConvLSTMPNet

To express the hydrodynamic memory effects, a multi-step forecasting model for roll motion requires a sequence-to-sequence architecture. Therefore, it is important to extract sequential correlation efficiently. In addition, the forecasting model for roll motions involves multiple variables as inputs, such as the time history of the roll angle and wave heights. Thus, it is also important to fuse the information from feature variables. Considering that LSTM and CNN have advantages in the above problems respectively, a parallel net of one-dimension CNN (Conv1D) and LSTM is designed, named ConvLSTMPNet.

A graphical illustration of the proposed model is shown in Fig. 2. The LSTM layer is used to learn the sequential correlation and the Conv1D layer is utilized to extract the local spatial-temporal information. They tackle the same inputs simultaneously. The outputs $C_t$ and $L_t$ of Conv1D and LSTM are connected end-to-end as inputs to the fully connected layer. Final, the fully connected layer provides multi-step prediction results of the roll motions. The flow of information is presented in Eq. (2)-(5):

$$C_t = g_1(W^c, [y_{t-d+1}, y_{t-d+2}, \ldots, y_t, X_t]) \qquad (2)$$

$$h_m = g_2(W^h, [y_n, x_n, h_{m-1}]) \qquad (3)$$

$$L_t = [h_1, h_2, \ldots, h_d] \qquad (4)$$

$$Y_t = (y_{t+1}, y_{t+2}, \ldots, y_{t+p}) = g_3(W^y, [C_t, L_t]) \qquad (5)$$

where $W^c$, $W^h$ and $W^y$ represents the weights of Conv1D, LSTM, and the fully connected layers, $g_1(\cdot)$, $g_2(\cdot)$, and $g_3(\cdot)$ represents nonlinear functions, $m = 1, 2, \ldots, d$; $n = t-d+1, t-d+2, \ldots, t$. The hidden state sequence $[h_1, h_2, \ldots, h_d]$ is treated as the temporal information in LSTM layer.

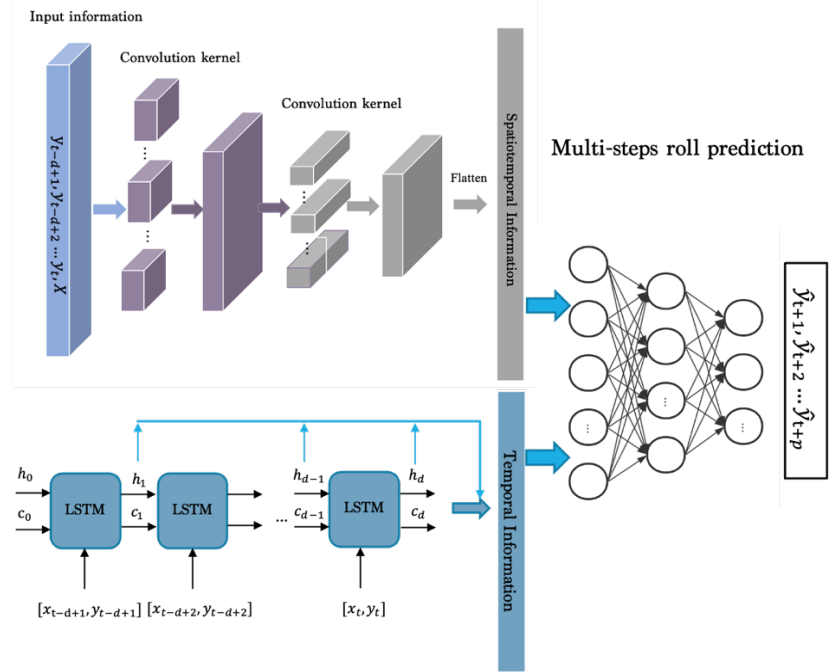

Fig. 2 Architecture of the ConvLSTMPNet

Specifically, LSTM is a variate RNN where the gate cell is introduced to solve the problem of gradient vanishing and gradient explosion in RNN (Hochreiter and Schmidhuber, 1997), as shown in Fig. 3. The forget gate $f_t$ decides what to forget from the previous memory cell $c_{t-1}$. The input gate $i_t$ controls what to read out of the candidate memory cell $\tilde{c}_t$ derived by state pair $[x_t, h_{t-1}]$, and $h_{t-1}$ is the previous hidden state. The output gate $o_t$ determines the value to be transferred into next training. Therefore, the attribute of transferring the previous hidden state into next cycle and above gates can help it better capture the long-term dependencies of roll time series.

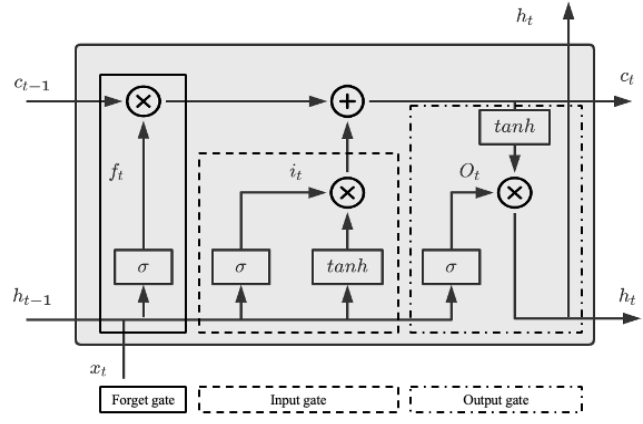

Fig. 3 Diagram of LSTM

The corresponding formulas of LSTM are given in Eq. (6) to Eq. (11):

$$f_t = \sigma(W^{f,h}h_{t-1} + W^{f,x}x_t + b_f) \tag{6}$$

$$i_t = \sigma(W^{i,h}h_{t-1} + W^{i,x}x_t + b_i) \tag{7}$$

$$o_t = \sigma(W^{o,h}h_{t-1} + W^{o,x}x_t + b_o) \tag{8}$$

$$\tilde{c}_t = tanh(W^{c,h}h_{t-1} + W^{c,x}x_t + b_c) \tag{9}$$

$$c_t = f_t \odot c_{t-1} + i_t \odot \tilde{c}_t \tag{10}$$

$$h_t = o_t \odot tanh(c_t) \tag{11}$$

where $W^{f,h}, W^{f,x}, W^{i,h}, W^{i,x}, W^{o,h}, W^{o,x}, b_f, b_i, b_o$ represent the weight matrixes and bias of forget gate, input gate and output gate, respectively. In addition, $W^{c,h}$ and $b_c$ denote the weight matrix and bias of the candiate cell state respectively; $\sigma$ and $\odot$ are logistic sigmoid function and elementwise multiplication, respectively.

In ConvLSTMPNet, as shown in Fig. 2, the entire hidden state sequences of LSTM are used as the output to provide more temporal information to the fully connected layer, which is specifically designed for multi-step prediction.

A typical Convolution neural network (CNN) contains convolutional layers, pooling layers and fully connected layers. The convolutional layers can extract features from input sequences. The convolution kernel of convolution layer features the parameter sharing, thus reducing the number of weights to be trained and the complexity of the network. The pooling layers can distill the extracted features and pay attention on the most salient elements. The fully connected layers can interpret the internal representation to implement task of regression or classification. Another important benefit of CNN is the support of multiple one-demensional (1D) inputs as a separate channel to extract key information, i.e., the Conv1D layer. One-demension convolution operation is expressed in the follwing equation:

$$c(t) = (x*\omega)(t) = \sum_{a} x(t+a)\omega(a) \quad (12)$$

where $x$ is the input features; $t$ demotes the time; $\omega$ represent weights; $a$ is the position of convolution kernel, and $c$ is the feature map output after convolution operation $*$.

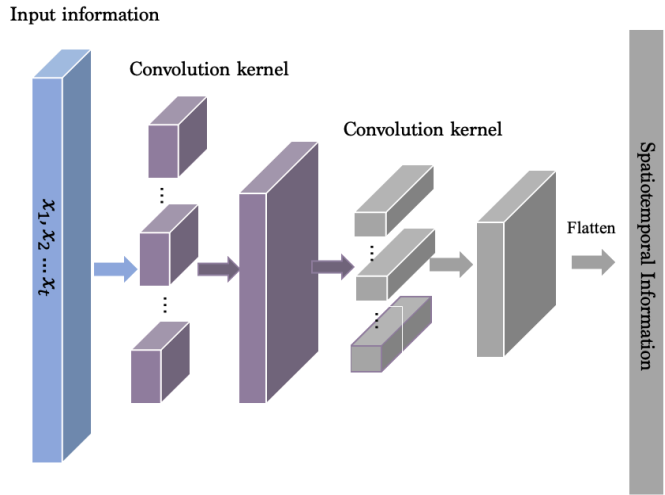

Fig. 4 The extraction of spatiotemporal information through 1D convolution

The CNN part in the ConvLSTMPNet is shown in Fig. 4. The 1D sliding convolution kernels move vertically to extract key spatiotemporal representation in local time span from past time series of roll and wave elevation. Moreover, different convolution kernels can extract feature from different perspectives to obtain higher-dimensional features. It is noted that no pooling layers are adopt due to small data size compared to image data.

Finally, the spatiotemporal and the time dependency information from CNN and LSTM will be flatten and concatenated into one vector, then the multi-step prediction of roll motion can be interpreted by the fully connected layers.

3.3 Feature selection

The selection of feature variables is critical to the accuracy of roll motion prediction, because proper feature variables can enhance the upper bound of the forecast model performance. Generally, two types of variables have been selected as features for the prediction of roll motions: motion states (Tang et al., 2021; Wei et al., 2021; Liu et al., 2020) and wave elevations (Del Águila Ferrandis et al., 2021; Xu et al., 2021). The relevant studies have demonstrated that both the roll motion states and wave elevation can provide some information for the prediction of roll motions. Compared with one-step ahead predicting, multi-step prediction has higher requirements for extracting sufficient information to guarantee the accuracy over a period of time.

Theoretically, if the data of motion states and wave heights involve different information, then the appropriate combination of these features can provide greater potential for forecasting modeling. However, few studies have investigated this problem. In this study, the time history of roll motion angle and the wave heights around ships are selects as feature variables. In Section 5.4, a comparative exploration of the feature selection is provided.

3.4 The generation of the numerical dataset

Datasets play a key role in model training and evaluation. However, available datasets under high sea state are scarce due to the high cost and navigational safety risks. In this study, a numerical technique based on computational fluid dynamics (CFD) is applied to obtain relatively high-fidelity data on ship motion in high sea states.

The datasets are generated by a Volume-of-Fluid (VOF) URANS solver in STAR-CCM+, which can resolve the viscous effects and nonlinear effects of ship motions. Specifically, the VOF method is used to model the free surface and imposes the velocity and pressure fields of superposition of the individual regular waves as the initial conditions in the domain to simulate the long crest irregular waves. Moreover, the quality of mesh is important for accurately simulating irregular wave, so the adaptive mesh method is adopted to generate mesh of free surface adaptatively. The overset mesh method which contains background domain and ship domain, and Dynamic Fluid Body Interaction (DFBI) model is used to simulate realistic ship motions, where pitch, heave and roll motions are collected.

Irregular long crested waves are simulated for the environmental disturbance. Assuming that waves is stationary, homogeneous and ergodic random process, the elevation of irregular wave is represented through surperposition of compoment regular wave as Eq. (13):

$$\zeta(t) = \sum_{i=1}^{n} a_i \cos(\omega_i t + \varepsilon_i) \qquad (13)$$

where the $\zeta$ denotes wave elevation, $\omega_i$ is frequency of ith component wave, $\varepsilon_i$ is random phases between $-\pi$ and $\pi$, and the component wave amplitude $a_i$ for a given frequency is obtained from the following equation:

$$\frac{1}{2} a_i^2 \cong S(\omega_i) \Delta \omega_i \qquad (14)$$

$$S_{PM}(\omega) = \frac{5}{16}(H_s^2 \omega_p^4)\omega^{-5} \exp(-\frac{5}{4}(\frac{\omega}{\omega_p})^{-4}) \qquad (15)$$

where $S_{PM}(\omega)$ is Pierson-Moskowitz (PM) spectrum, $H_s$ is significant wave elevation and $\omega_p = (2\pi/T_p)$ represents the angular peak frequency.

## 4 Data Preparation and Validation

### 4.1 Data Generation

The KCS ship is taken as the study object. Considering the computational cost, the scale KCS ship model is adpoted. The dimensions of the full scale and model scale KCS are given in TABLE 1 and the geometry of KCS is showed in Fig. 5. The extreme sea condition is selected as sea state 7. The Superposition of Wave method in VOF module is adopted to generate irregular long crest wave following PM spectrum with 240 linear regular component waves at equal frequency interval. Referring to the World Regulation of the State Oceanic Administration (TABLE 2), the specific wave parameters for the scale model in sea state 7 are set as follows: the significant wave height $H_s = 0.2215m$ (corresponding 7m in full scale) and the peak wave period $T_p = 2.16s$ (corresponding 12.13s in full scale).

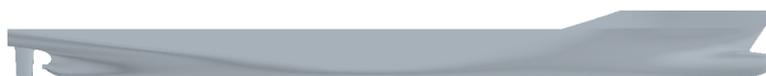

Fig. 5 KCS hull model in STAR-CCM+

TABLE 1 Dimension of KCS

| Dimensions | Full Scale | Model Scale |
|---|---|---|
| Scale | 1 | 31.599 |
| Length between the perpendiculars(m) | 230 | 7.2786 |
| Beam at waterline(m) | 32.2 | 1.0190 |
| Depth(m) | 19 | 0.6013 |
| Draft(m) | 10.8 | 0.3418 |
| Displacement (m³) | 52030 | 1.6490 |
| Longitudinal center of gravity from the aft(m) | 111.6 | 3.532 |
| Inertia moment of x axis / Beam at waterline | 0.4 | 0.40 |
| Inertia moment of z axis / Length between the perpendiculars | 0.26 | 0.26 |
| Design speed (m/s, full scale: kn) | 24 | 2.196 |
| Froude number | 0.26 | 0.26 |

TABLE 2 Classifications of wave level

| Sea state | Significant wave elevation |
|---|---|
| 1 | <0.1 |
| 2 | 0.1~0.5 |
| 3 | 0.5~1.25 |
| 4 | 1.25~2.5 |
| 5 | 2.5~4.0 |
| 6 | 4.0~6.0 |
| 7 | 6.0~9.0 |
| 8 | 9.0~14.0 |
| 9 | >14.0 |

Numerical datasets are constructed for three conditions with corresponding wave directions of 150°, 120°, and 90°, called Dataset #1, Dataset #2, Dataset #3, respectively. It is noted that the wave direction at 90° corresponds to the port beam wave, 0° to the following wave and 180°to the head wave. The total number of grid points is around 1,470,000. The ship motions are calculated over 80s for each case. The time step is set to 0.005s to satisfy the Courant-Friedrichs-Lewy condition. The time series of roll angle and the wave height around the hull at three points are collected for training. In the CFD numerical tank, three wave gauges is arranged around the hull to monitor the free surface elevation as shown in Fig. 6 (a). Gauge 1 is 0.2 times of ship length away from the bow, and Gauge 2 and 3 are 1.5 times of ship width from the hull to detect the encounters wave around the ship. The snapshot of simulation is shown in Fig. 6 (b).

The roll motion response exhibits different characteristics when the hull encounters waves from different directions (as shown in Fig. 7). Specifically, a) when the ship encounters 90° beam wave, the amplitude of roll angles is the largest, up to 30°. b) when the ship encounters the oblique waves of 120° and 150°, the roll angles stand in [-20°,15°] and [-10°, 8°], respectively. However, the nonlinear characteristics of the roll motion under oblique waves are more significant than those under beam waves. Theoretically, when the ship encounters beam wave, the encounter frequency is equal to the wave frequency. In the conditions of 120° and 150° oblique waves, the encounter frequency is larger than the wave frequency, leading to a higher frequency response than that of beam waves. Therefore, the theoretical analyzes are consistent with the simulation results.

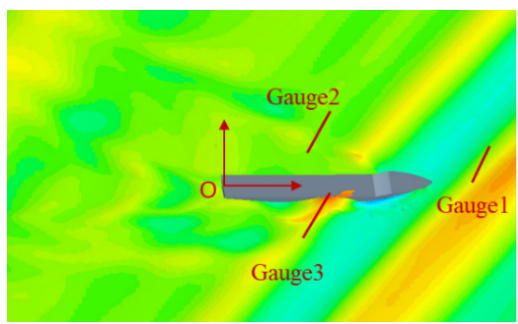

(a) The location of wave gauge near the hull.

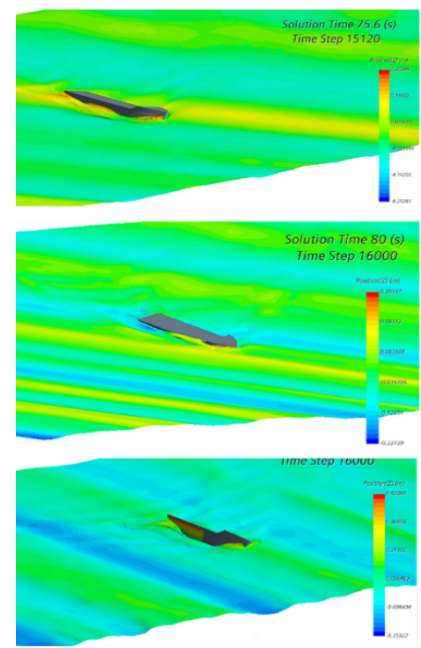

(b) From top to bottom, the wave heading angles are 150°, 120° and 90°, respectively

Fig. 6 Snapshots of simulation.

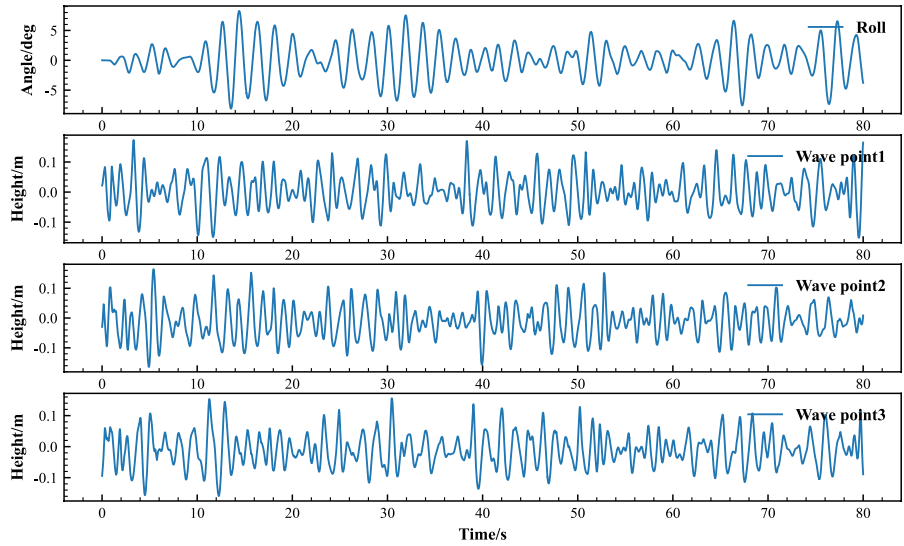

(a) Dataset #1: the wave heading angles of 150°

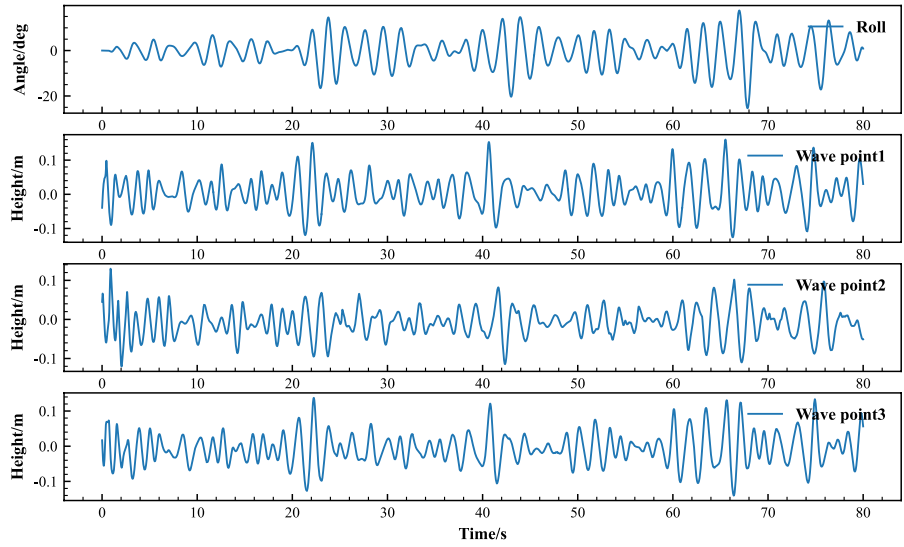

(b) Dataset #2: the wave heading angles of 120°

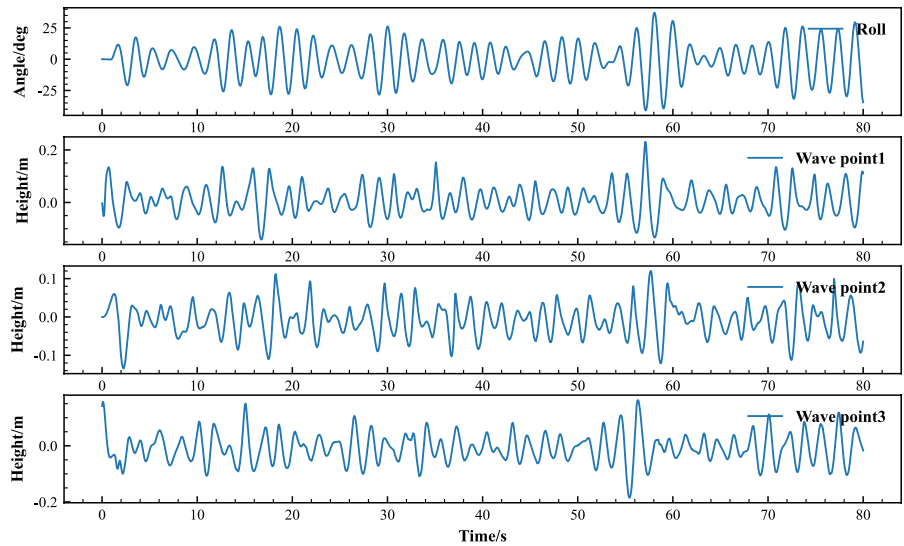

(c) Dataset #3: the wave heading angles of 90°

Fig. 7 Three simulated datasets in sea state 7

## 4.2 Validation of numerical datasets

The validation of the input irregular wave refers to the literature (Diez et al. 2018). The irregular wave generated by CFD is validated versus theoretical benchmark values from the spectrum. In Fig. 8, the CFD spectrum are compared to the nominal spectrum. In addition, the wave energy spectrum moments $m_0$, $m_1$, and $m_2$ are applied as additional variables to evaluate the accuracy of the CFD wave energy spectrum, as Eqs. (16-17):

$$m_k = \int_0^\infty \omega^k S(\omega) d\omega \quad k = 0,1,2 \tag{16}$$

$$E = m_k^{(CFD)} - m_k^{(t)} \tag{17}$$

where superscripts CFD represents simulation value and $t$ represents theoretical values, $E$ indicates the error associated with the CFD wave energy spectrum moments.

The wave energy spectral moments for CFD and theoretical are presented in TABLE 3. The average errors $E$ are small with an average of 0.026%. The results demonstrate that the CFD wave energy spectrum achieves good agreement with theorical values, verifying the effectiveness of input irregular waves generated from CFD method.

TABLE 3 Wave energy spectrum moments of CFD and theorical results

| Analysis CFD vs Theory | N.wave Comp. per run | Tot. run time/ $T_p$ | $m_0$ | | $m_1$ | | $m_2$ | |
|---|---|---|---|---|---|---|---|---|
| | | | Value [m²] | E% | Value [m²]rad/s | E% | Value [m²] rad²/s² | E% |
| Inputwave | 240 | 37 | 0.003055 | -0.00061% | 0.01149 | 0.0048% | 0.049561 | 0.0715% |

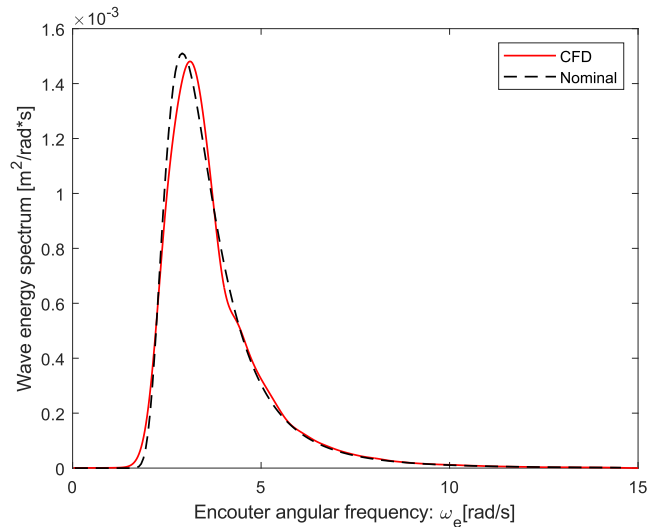

Fig. 8 Wave energy spectra of CFD results versus nominal results

To test the availability of numerical results for the interaction between ships and waves, the simulated KCS motions in regular waves are compared with the experiment data from 2015 Tokyo Workshop (T2015 Workshop). Fig. 9 shows the comparison results for the wave length-ship length ratio $\lambda / L = 1.15, 1.37$, the wave height-ship length ratio $H/L = 1/60$ and Froude number $Fr = 0.26$. The available data on heave, pitch and wave elevation are compared and the simulated results make a great agreement with experimental data.

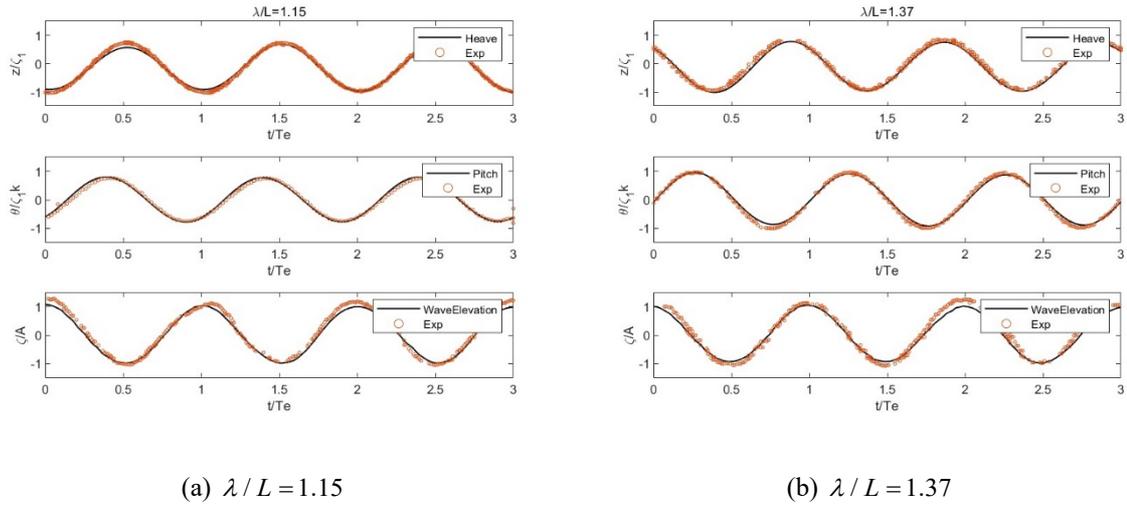

(a) $\lambda/L = 1.15$  (b) $\lambda/L = 1.37$

Fig. 9 Validation case 2: Response of KCS under regular wave

## 5 Study Case: multi-step prediction of the roll motion in sea state 7

5.1 Data processing

For multi-step prediction of ship roll motions, the dataset is divided into a training set and a validation set with a ratio of 8:2. The data are normalized in the interval (0, 1) to improve the convergence of the model. Furthermore, the structure of the training data for each time step is translated into a sliding window form, such as $\left[x_{t-d+1}, y_{t-d+1}, \ldots, x_t, y_t, \ldots, y_{t+p}\right], \left[x_{t-d+2}, y_{t-d+2}, \ldots, x_{t+1}, y_{t+1}, y_{t+2}, \ldots, y_{t+p+1}\right]$ (as shown in Fig. 10). The green boxes represent the input sequence, and the pink boxes represent the output sequence that needs to be predicted.

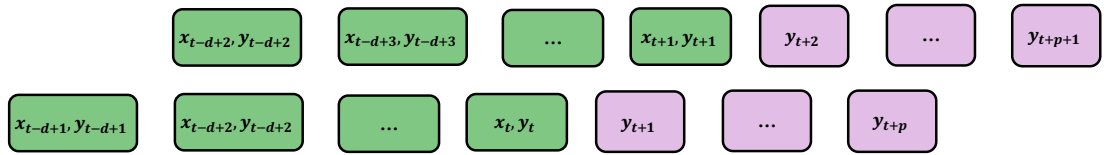

Fig. 10 Sliding window method

5.2 Model setting

To optimize the hyperparameters of the proposed algorithm, the cross-validation on a rolling basis is conducted. The specific mode is shown in the Fig. 11. The original data for training is divided into k blocks. At first turn, several blocks of data are used for training and one block is used for testing. Then the current test block is concatenated as part of the training data for the next round. This process continues multiple times until the data is fully utilized. Finally, the average mean square error is used to evaluate the performance of the hyperparameter set.

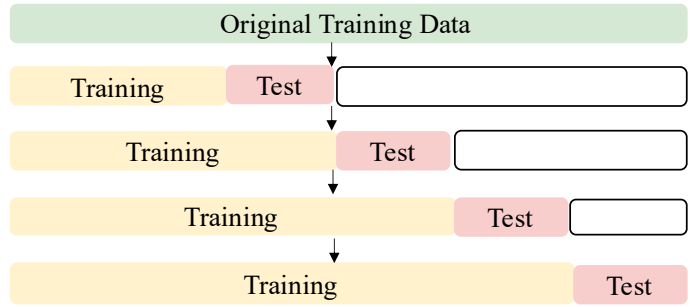

Fig. 11 Cross-validation on a rolling basis for time series prediction

The ConvLSTMPNet architecture contains a LSTM layer and two Conv1D layers. The number of convolution filters and neural units of hidden layer in LSTM and CNN are investigated by the cross-validation on a rolling basis. The hyperparameter candidates of the convolution filter are set to [32, 64, 128] with fixed kernel size of 3, and the neural units of the hidden layer in the LSTM are set to [32,64,100,128]. The parameters of three fully connected layers in ConvLSTMPNet are fixed which are 100, 50 and the same number of units as the steps desired to be predicted. Based on the method of cross-validation on a rolling basis, the relatively optimal parameters of above three models are obtained and listed in TABLE 4 and Fig. 12. Then these parameters are used to conduct following case study in Section 5.4 and 5.5.

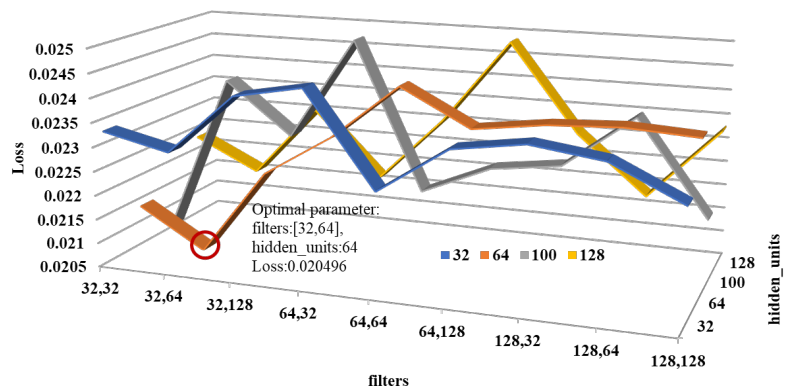

(a) Cross validation of parameters of ConvLSTMPNet

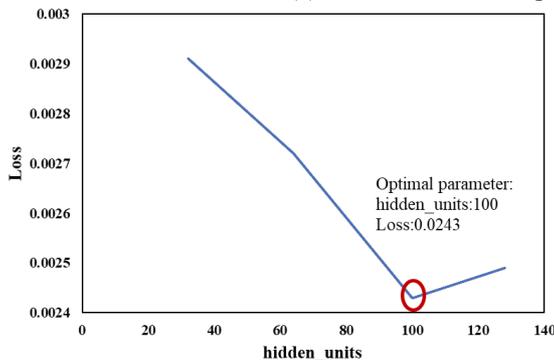

(b) Cross validation of parameters of LSTM

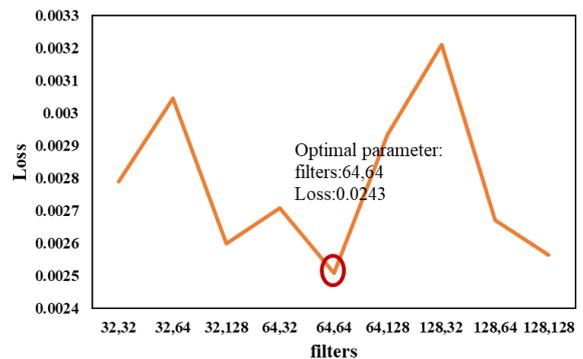

(c) Cross validation of parameters of CNN

Fig. 12 Cross validation of parameters of CNN, LSTM, ConvLSTMNet

TABLE 4 Cross validation of optimal parameters on LSTM, CNN, ConvLSTMPNet

| Candidate parameters | Model | | |
|---|---|---|---|
| | ConvLSTMPNet | LSTM | CNN |
| filters: [32,64,128] | 32,64 | None | 64,64 |
| hidden_units: [32,64,100,128] | 64 | 100 | None |

Specifically, the ConvLSTMPNet architecture contains one LSTM with 64 units in the hidden layer and two Conv1D layers with 32 and 64 filters, with a kernel size of 3. The outputs of CNN and LSTM are concentrated together as the input for the three fully connected layers to generate the final prediction. In the three fully connected layers, the first and second layers have 100, 50 units, respectively, and the third layer has the same number of units as the steps desired to be predicted. Theoretically, the ConvLSTMPNet synthesizes the advantages of the two models. To evaluate its performance, LSTM and CNN are selected as the comparison model. The LSTM model has 100 units in its hidden layer, as same as the setting of fully connected layers in ConvLSTMPNet model. For CNN model, two Conv1D layers with 64 filters and kernel sizes of 3, the flatten results are transferred into the fully connected layers which has the same setting as the ConvLSTMPNet model to predict the roll motions. The length of the lag time is set to be equal to the length of the time to be predicted. The batch size is set to 50 and the above model is optimized by Adam optimizer.

5.3 Model Evaluation

For the evaluation of the model, it should be mentioned that the main objective to be achieved is prediction, rather than explanation. This is because the data-driven method has the advantage of mapping nonlinear features through supervised learning, but somehow lacks explainability. Therefore, the results are mainly analyzed and evaluated from the perspective of the prediction effects by the index of RMSE and the visualization of the prediction performance. Mean Square Error (MSE) is taken as the loss function for training, and Root Mean Square Error (RMSE) is taken to evaluate the performance of the predictive model. The formulas are given in Eq. (18) and Eq. (19), where n is the total number of samples, $y_i$ denotes the real value and $\hat{y}$ denotes the predict value.

$$MSE = \frac{\sum_1^n (y_i - \hat{y}_i)^2}{n} \tag{18}$$

$$RMSE = \sqrt{\frac{\sum_1^n (y_i - \hat{y}_i)^2}{n}} \tag{19}$$

5.4 Study Case I: Evaluating the feature space for the multi-step prediction of roll motions

Theoretically, the roll motion is dependent to the time history of the motion states due to the hydrodynamic memory effect. On the other hand, the wave elevation around the ship has been proven to be effective for roll motion prediction because it directly reflects the wave effects. However, the impact of these variables on the prediction performance remains to be studied to figure out how to extract the maximum information. Therefore, this study case considers both the time history of motion states and wave elevation as the feature space, and progressively investigates the impact of the following three scenarios on multi-step prediction: (1) only roll angle, (2) only wave elevation, and (3) both roll angle and wave elevation. The lengths of the time to be predicted are chosen to be 10 and 20 steps which corresponds to covering at least one period of roll motion (around 11s for full scale). The learning method applied here is the conventional LSTM algorithm, described in Section 5.2.

The average RMSEs of 10- and 20-step predictions are listed in TABLE 5. Figs. 13-14 show the prediction results under different feature space. Fig. 13 presents the assessment of the predictive accuracy for each time step under different feature spaces, represented by RMSE. Fig. 14 shows the comparison results between the predicted and actual values for roll motion under different input features space, where the black line indicates actual value, and the dashed line denotes predicted values under differnet feature variables. The left column shows the prediction results for 10 steps, and the right column shows the prediction results for 20 steps.

TABLE 5 Average RMSE of 10 steps and 20 steps roll prediction

| Timesteps | 10 steps | | | 20 steps | | |
| --- | --- | --- | --- | --- | --- | --- |
| Feature space | Roll angle | Wave elevation | Roll angle & Wave elevation | Roll angle | Wave elevation | Roll angle & Wave elevation |
| Dataset #1 | 1.17956 | 1.53057 | **0.69255** | 1.55123 | 1.51943 | **0.92316** |
| Dataset #2 | 3.47372 | 3.58599 | **2.69369** | 5.15763 | 4.15442 | **3.43608** |
| Dataset #3 | 3.03476 | 4.95515 | **1.20959** | 4.45659 | 2.79653 | **1.46759** |

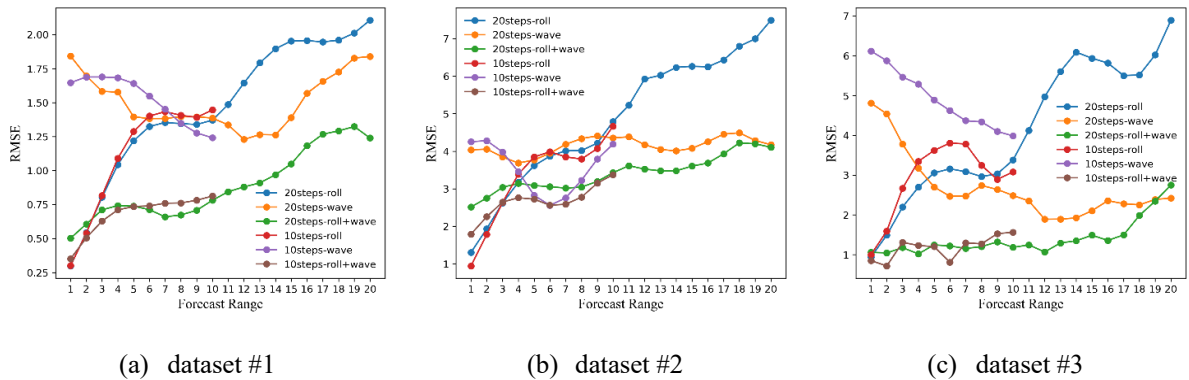

(a) dataset #1  (b) dataset #2  (c) dataset #3

Fig. 13 The RMSE of each step in multi-step prediction

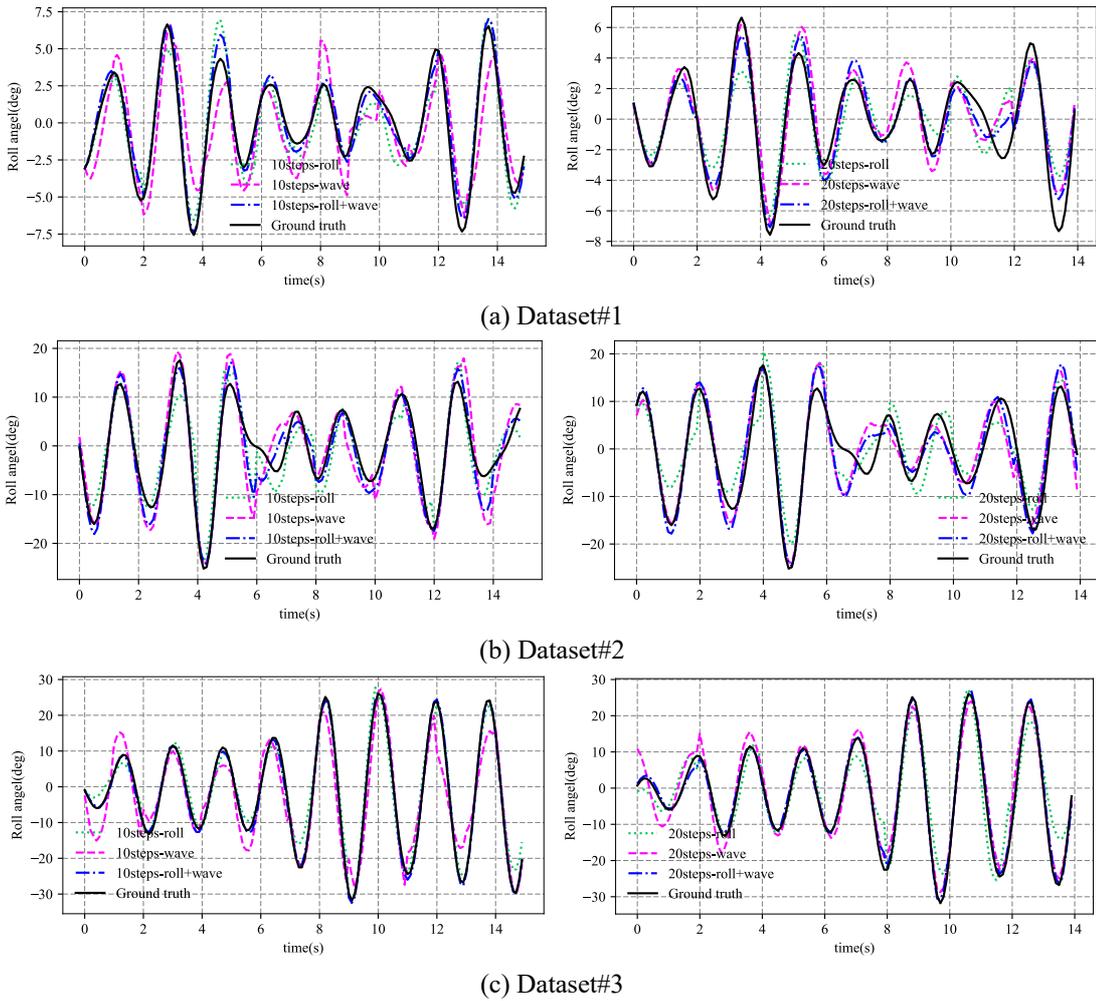

(a) Dataset#1

(b) Dataset#2

(c) Dataset#3

Fig. 14 The comparison results with different feature variables under dataset #1, #2 and #3

From Fig. 14, the roll and wave elevation can provide some valid information for predicting the roll motions, respectively. However, in the case of roll motion as the feature space, underestimation of large amplitude occurs due to the lack of external wave disturbance. Overall, it can be seen from Fig. 13 that when the roll motion states are used as the feature variable, the prediction error increases with the number of steps. Whereas, when wave heights are used as the feature variable, the error decreases with increasing number of steps. This result indicates that the information contained in the roll motion state may be limited compared to the wave height, while wave height requires more steps as input to reflect the effect of waves on roll motions. The results in Fig. 13, Fig. 14 and Table 5 indicate that the feature space containing wave elevation and roll motion states can provide more sufficient information than using only one of them. With the development of the wave radar technology, it is recommended to use both wave elevation and roll motion state as feature variables to enhance the prediction of roll motions in extreme sea conditions.

5.5 Study Case II: Multi-steps roll motion prediction based on ConvLSTMPNet, CNN and LSTM

The effects of the learning algorithm are investigated in this section under the premise of selecting both the time history of motion states and wave elevation information as the feature space. As seen in Study Case 1, the 20-step prediction in oblique waves (for dataset #1 and #2) still has room for improvement as the more complicate coupling between the ship roll motion and the waves. To enhance the extraction efficiency of multidimensional information, the proposed ConvLSTMPNet is applied and evaluated by comparing it with LSTM and CNN described in Section 5.2.

The average RMSEs of the dataset #1 and #2 are listed in TABLE 6 and the RMSE of each step in the multi-steps prediction is shown in Fig. 15. It can be found that ConvLSTMPNet has the best performance, with the lowest average RMSE of 0.80294 and 3.22126 under dataset #1, # 2, respectively. Additionally, for each step of the 20-step prediction, ConvLSTMPNet obtains the highest predictive accuracy. In contrast, the performance of LSTM is slightly worse than that of the proposed hybrid model, but better than that of CNN.

TABLE 6 Average RMSE of CNN, LSTM and ConvLSTMPNet

| Dataset | #1 | | | #2 | | |
| --- | --- | --- | --- | --- | --- | --- |
| Model | CNN | LSTM | ConvLSTMPNet | CNN | LSTM | ConvLSTMPNet |
| Average RMSE | 1.05856 | 0.92316 | **0.80294** | 3.4442 | 3.43608 | **3.22126** |

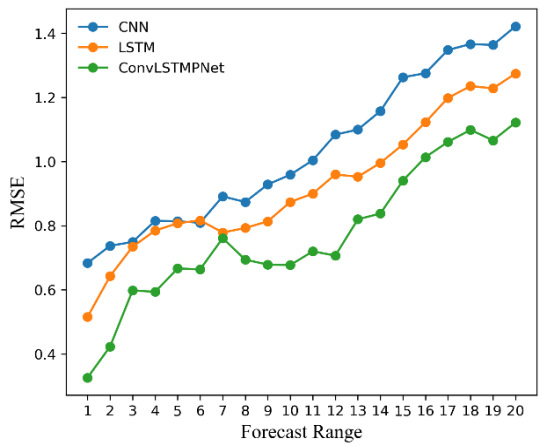
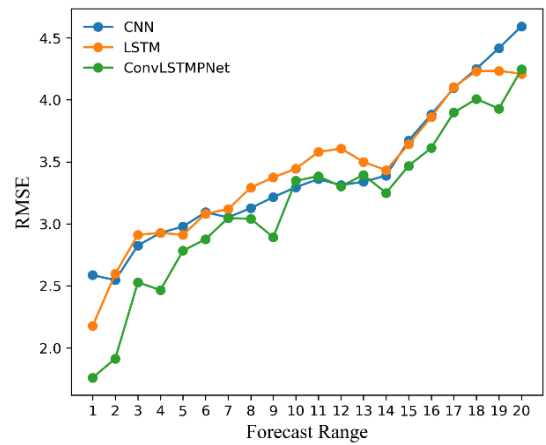

(a) dataset #1   (b) dataset #2

Fig. 15 The RMSE of each step in multi-steps prediction: (a) dataset #1; (b) dataset #2

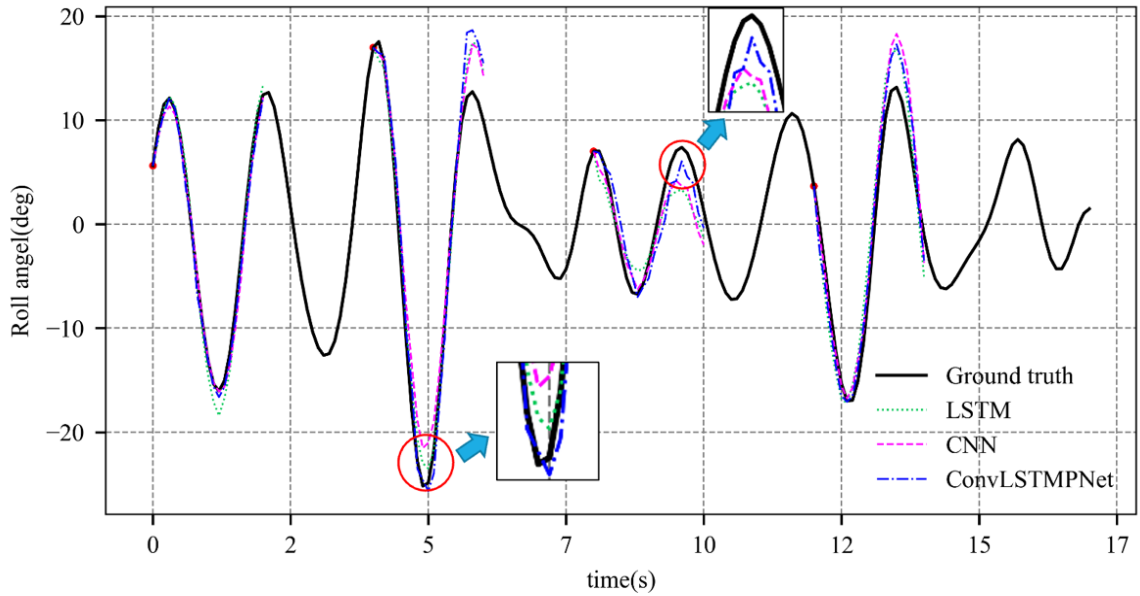

Fig. 16 The prediction results with CNN, LSTM and ConvLSTMPNet for dataset #2

In Fig. 16, four points are selected at intervals for dataset #2 and each of the 20-step prediction are conducted, which is 2s for the scaled model and 11s for the full scale. The results demonstrate that at least one period of roll motion can be accurately predicted by using the proposed method. Compared with the single LSTM and CNN method, the proposed method has better performance in the prediction of the amplitude of roll angles. From the perspective of architectural composition, the ConvLSTMPNet architecture combines the advantages of LSTM and CNN in parallel to extract time memory effects and nonlinear coupling interaction between incident wave and ship motion, resulting in more accurate multi-step prediction results. In summary, the results demonstrate the effectiveness of the proposed method for the multi-step prediction of roll motion in high sea states.

## 6 Conclusion and future work

In this paper, a data-driven methodology is proposed for multi-step prediction of ship roll motion in high sea states. A hybrid neural network ConvLSTMPNet which combines the LSTM and Conv1D in parallel is exploited to extract the nonlinear dynamics characteristics and the hydrodynamic memory information from wave and roll motion states, so as to obtain accurate multi-step predictions. The numerical solutions of KCS in sea state 7 irregular long-peaked waves with different wave directions are generated as datasets by CFD method. The comparative study of feature selection demonstrates the superiority of selecting both motion states and wave heights as the feature space. The proposed method can accurately predict at least one period of roll motion in high sea state. The accurate prediction of roll motion can benefit the operation and safety of marine vessels and support the development of decision-making technologies for autonomous ships.

## 7 Acknowledgment

This work is financially supported by the National Natural Science Foundations of China [Grant number: 61973208, 61991415, U1813217 and 52101361], the "Shuguang Program" 20SG40 supported by Shanghai Education Development Foundation and Shanghai Municipal Education Commission and the program of shanghai academic research leader 20XD1421700.

# 8 Reference

Bassler, C.C., 2013. Analysis and modeling of hydrodynamic components for ship roll motion in heavy weather (PhD Thesis). Virginia Polytechnic Institute and State University.

Bengio, Y., Simard, P., Frasconi, P., 1994. Learning long-term dependencies with gradient descent is difficult. IEEE transactions on neural networks / a publication of the IEEE Neural Networks Council 5, 157–66.

Bergamasco, F., Benetazzo, A., Yoo, J., Torsello, A., Barbariol, F., Jeong, J.-Y., Shim, J.-S., Cavaleri, L., 2021. Toward real-time optical estimation of ocean waves' space-time fields. Computers & Geosciences 147, 104666.

Cang, Y., He, H., Qiao, Y., 2019. Measuring the Wave Height Based on Binocular Cameras. Sensors (Basel) 19, 1338.

Chung, J.-S., Bernitsas, M.M., 1997. Hydrodynamic Memory Effect on Stability, Bifurcation, and Chaos of Two-Point Mooring Systems. Journal of Ship Research 41, 26–44.

Del Águila Ferrandis, J., Triantafyllou, M.S., Chryssostomidis, C., Karniadakis, G.E., 2021. Learning functionals via LSTM neural networks for predicting vessel dynamics in extreme sea states. Proceedings of the Royal Society A 477, 20190897.

D'Agostino, D., Serani, A., Stern, F., Diez, M., 2021. Recurrent-type neural networks for real-time short-term prediction of ship motions in high sea state. arXiv preprint arXiv:2105.13102.

Diez, M., Broglia, R., Durante, D., Olivieri, A., Campana, E.F., Stern, F., 2018. Statistical assessment and validation of experimental and computational ship response in irregular waves. Journal of Verification, Validation and Uncertainty Quantification 3.

Ferrandis, J. del Á., Triantafyllou, M., Chryssostomidis, C., Karniadakis, G., 2019. Learning functionals via LSTM neural networks for predicting vessel dynamics in extreme sea states. arXiv:1912.13382.

Hochreiter, S., Schmidhuber, J., 1997. Long Short-term Memory. Neural computation 9, 1735–80.

Huang, B. G., Zou, Z., Ding, W., 2018. Online prediction of ship roll motion based on a coarse and fine tuning fixed grid wavelet network. Ocean Engineering 160, 425–437.

Hou, X.-R., Zou, Z.-J., 2016. Parameter identification of nonlinear roll motion equation for floating structures in irregular waves. Applied Ocean Research 55, 66–75.

Inoue, S., Hirano, M., Kijima, K., Takashina, J., 1981. A practical calculation method of ship maneuvering motion. International Shipbuilding Progress 28, 207–222.

Jiang, H., Duan, S., Huang, L., Han, Y., Yang, H., Ma, Q., 2020. Scale effects in AR model real-time ship motion prediction. Ocean Engineering 203, 107202.

Jiang, Y., Wang, X.-G., Zou, Z.-J., Yang, Z.-L., 2021. Identification of coupled response models for ship steering and roll motion using support vector machines. Applied Ocean Research 110, 102607.

Jiao, J., Chen, C., Ren, H., 2019. A comprehensive study on ship motion and load responses in short-crested irregular waves. International Journal of Naval Architecture and Ocean Engineering 11, 364–379.

Li, J., 2003. Ship Seakeeping Performance.

Liu, Y., Duan, W., Huang, L., Duan, S., Ma, X., 2020. The input vector space optimization for LSTM deep learning model in real-time prediction of ship motions. Ocean Engineering 213, 107681.

Lyzenga, D.R., Nwogu, O.G., Beck, R.F., O'Brien, A., Johnson, J., de Paolo, T., Terrill, E., 2015. Real-time estimation of ocean wave fields from marine radar data, in: 2015 IEEE International Geoscience and Remote Sensing Symposium (IGARSS). pp. 3622–3625.

Newman, J.N., 2018. Marine hydrodynamics. The MIT Press.

Serani, A., Diez, M., Walree, F. van, Stern, F., 2021. URANS analysis of a free-running destroyer sailing in irregular stern-quartering waves at sea state 7. Ocean Engineering 237, 109600.

Sutskever, I., Vinyals, O., Le, Q.V., 2014. Sequence to Sequence Learning with Neural Networks. arXiv:1409.3215.

Sun, Q., Tang, Z., Gao, J., Zhang, G., 2022. Short-term ship motion attitude prediction based on LSTM and GPR. Applied Ocean Research 118, 102927.

Tang, G., Lei, J., Shao, C., Hu, X., Cao, W., Men, S., 2021. Short-Term Prediction in Vessel Heave Motion Based on Improved LSTM Model. IEEE Access 9, 58067–58078.

T2015 Workshop. [online] Available at: http://www.t2015.nmri.go.jp [Accessed 17 Nov. 2022]

Wang, F., Yuan, G., Lu, Z., 2007. Investigation of Real-Time Wave Height Measurement Using X-Band Navigation Radar, in: 2007 International Conference on Wireless Communications, Networking and Mobile Computing. IEEE, Shanghai, China, pp. 980–983.

Wang, J., Zou, L., Wan, D., 2017. CFD simulations of free running ship under course keeping control. Ocean Engineering 141, 450–464.

Wei, Y., Chen, Z., Zhao, C., Tu, Y., Chen, X., Yang, R., 2021. A BiLSTM hybrid model for ship roll multi-step forecasting based on decomposition and hyperparameter optimization. Ocean Engineering 242, 110138.

Xu, W., Maki, K.J., Silva, K.M., 2021. A data-driven model for nonlinear marine dynamics. Ocean Engineering 236, 109469.

Yasukawa, H., Yoshimura, Y., 2015. Introduction of MMG standard method for ship maneuvering predictions. Journal of Marine Science and Technology 20, 37–52.

Yin, J., Zou, Z., Xu, F., 2013. On-line prediction of ship roll motion during maneuvering using sequential learning RBF neuralnetworks. Ocean Engineering 61, 139–147.

Yumori, I., 1981. Real Time Prediction of Ship Response to Ocean Waves Using Time Series Analysis, in: OCEANS 81. Presented at the OCEANS 81, pp. 1082–1089.

Zhang, W., Wu, P., Peng, Y., Liu, D., 2019. Roll motion prediction of unmanned surface vehicle based on coupled CNN and LSTM. Future Internet 11, 243.